\documentclass[10pt,twocolumn,letterpaper]{article}

\usepackage{times}
\usepackage{epsfig}
\usepackage{graphicx}
\usepackage{amsmath}
\usepackage{amssymb}
\usepackage[letterpaper]{geometry}

\usepackage[breaklinks=true,bookmarks=false]{hyperref}

\date{\vspace{-5ex}}
\begin{document}

\title{Efficient Two-Stream Motion and Appearance 3D CNNs for \\Video Classification}

\author{Ali Diba \\
ESAT-KU Leuven \\
{\tt\small ali.diba@esat.kuleuven.be}\and Ali Pazandeh \\
Sharif UTech \\
{\tt\small pazandeh@ee.sharif.ir}
\and Luc Van Gool\\
ESAT-KU Leuven, ETH Zurich \\
{\tt\small luc.vangool@esat.kuleuven.be}
}
\maketitle

\begin{abstract}
The video and action classification have extremely evolved by deep neural networks specially with two stream CNN using RGB and optical flow as inputs and they present outstanding performance in terms of video analysis. One of the shortcoming of these methods is handling motion information extraction which is done out side of the CNNs and relatively time consuming also on GPUs. So proposing end-to-end methods which are exploring to learn motion representation, like 3D-CNN can achieve faster and accurate performance.
We present some novel deep CNNs using 3D architecture to model actions and motion representation in
an efficient way to be accurate and also as fast as real-time. Our new networks learn distinctive models to combine deep motion features into appearance model via learning optical flow features inside the network.

\end{abstract}

\section{Introduction}

Recent efforts on human action recognition focused on using the spatio-temporal information of the video as efficient as possible \cite{2stream,TDD}. To do so there are many different view points to the problem. Considering the video as a 3D Volume or a sequence of 2D frames are the most common ones \cite{c3d,TDD}. Despite the promising results of convolutional networks on most of the visual recognition tasks, having temporal information, and fine-grained classes exclude action recognition task from other recognition task in having a significant gap between the results of deep and handcrafted features based methods. The two base factors of the case are first, having not enough number of videos in proposed datasets and second, disability of the proposed works in handling the temporal information as well as other tasks \cite{2stream-cvpr16}. As it can be inferred from the results which reported in number of the works, the network which works on optical flow have more discrimination on action classes. Also it needs a pre-process for each set of frames to compute the optical flow which is time consuming. In the other hand, the works which does not compute the flow and try to handle the temporal information by the network like C3D\cite{c3d} have less accuracy than two stream works. We believe that the main cause of this occurrence is insufficiency of training data. So the network can not be learnt to handle the temporal information of the video, considering training phase is from the scratch due to the differences of the proposed network to common networks.   

\begin{figure}[t]
 \centering
\includegraphics[width=250pt]{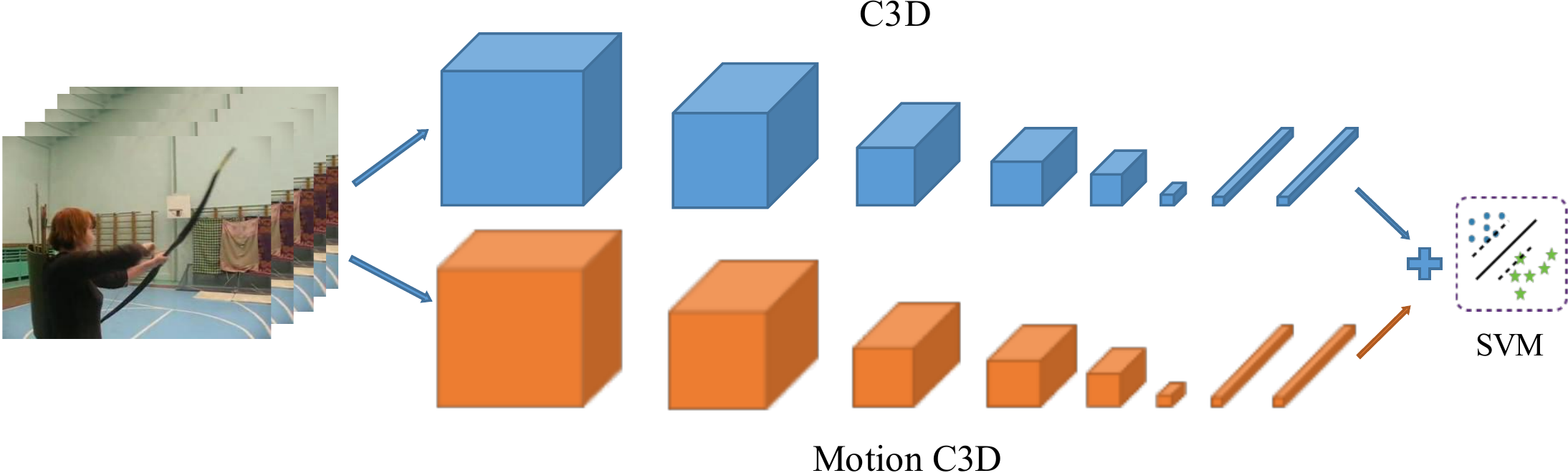}
 \caption{We use 3D-Convnet to learn motion from optical flow and extract the mid-level features to combine with regular C3D features as a new representations for videos. This figure shows our initial idea to combine these features.}
 \label{fig:1}
 \end{figure}

In conclusion we need a network with power of two-stream networks in handling temporal information beside the performance of C3D in time and handling spatial data. To reach this goal we proposed our two-stream 3D network which use 3D convolutions in both spatial and temporal streams. Also it uses the abstract feature vector of the optical flow estimation network. This abstract feature is obtained from the last layer of the convolution part in the flow estimation network.

In the fallowing, we discuss related works in section \ref{Related}, in section \ref{Method} we describe our proposed networks. Finally in section \ref{Experiments} the results of the proposed networks reported on the common datasets.

\section{Related works}
\label{Related}
Most of the recent works on visual recognition tasks and specially human action recognition based their works on using convolutional neural networks to perform better than previous works. As usually action recognition datasets contain videos, proposed methods effort to benefit from the temporal information beside the spatial information of each frame.

 \begin{figure}[ht]
 \centering
 \includegraphics[width=200pt]{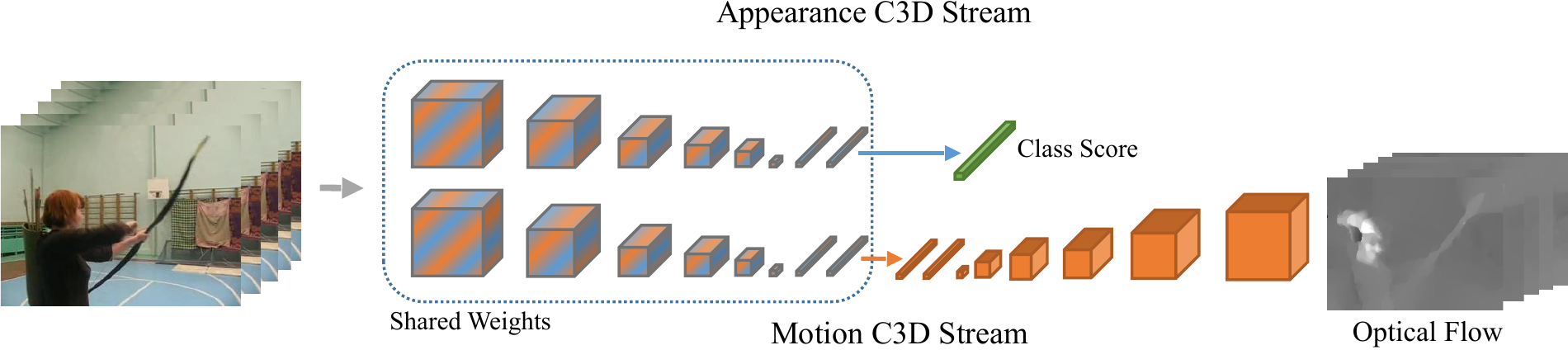}
 \caption{Our second idea to train an end-to-end 3D CNN with two loss function to classify video and estimate the optical flow. The 3D-Convnet part is shared for the two tasks.}
 \label{fig:2}
 \end{figure}

The two stream architecture \cite{2stream} proposed by simonyan et al. extracts the temporal data by training a CNN alexnet network on the optical flow which computed between consecutive frames. Many works offer to extend the proposed idea of two stream network by different views. Wang et al.\cite{verydeep} tried to improve the result by using deeper networks. Gkioxari et al.\cite{actiontubes} proposed an action detection method based on the two stream network. Feichtenhofer et al. \cite{2stream-cvpr16} extended the two stream network by implementing different fusion methods in different layers instead of the late fusion in the score layer of the two stream network of \cite{2stream}. As they claimed in their work, in contrast with most of the works, their results got state of the art without combining with IDT\cite{IDT} method.

Donahue et al.\cite{LongRecurrent} handled the temporal information by using a long short term memory on the extracted features of frames. Duo to not having an end-to-end network, the results have not improve as much as expected. 

The handcrafted features of improved trajectory proposed by wang et al.\cite{IDT} have considerable results as well as having the power of improving convolutional neural network based results in combination with them. Hence most of the proposed works got advantage of this power by concatenating the IDT feature with the proposed feature. Wang et al.\cite{TDD} use the power of IDT by extracting convolutional neural network based features locally around the trajectory and then encode the local features by fisher vector encoding.

As it can be inferred from the reviewed methods, most of them use either optical flow or handcrafted features (IDT) to improve the performance. In one hand both of the ideas are time consuming, and in the other a network exist which can perform better than the handcrafted features or inputs. In the following we describe our proposed network on raw images, which handles the temporal information, despite using a network architecture which is trained on the Imagenet dataset. 

 \begin{figure*}[ht]
 \centering
 \includegraphics[width=500pt]{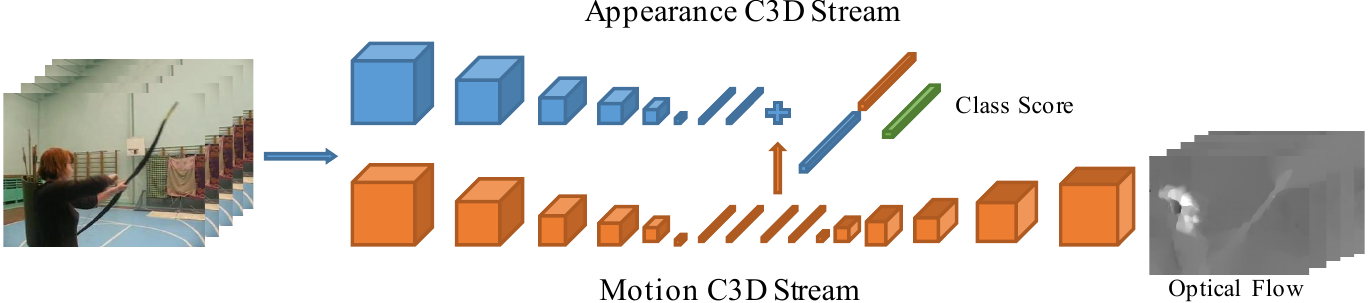}
 \caption{The proposed two stream end-to-end 3D CNN learning for video classification.}
 \label{fig:3}
 \end{figure*} 

\section{Learning Appearance and Motion CNNs}
\label{Method}
We will explain in detail the concepts of our proposed methods and analyze different architectures for 3D ConvNets of appearance and motion to achieve better performing CNNs for action and video classification empirically, and a training scheme on large scale video datasets for feature learning.

Inspired by the recently proposed C3D method \cite{c3d,dc3d} to learn 3d convolution and deconvolution networks, we designed our networks to learn a new feature representation of videos by exploiting a novel way of training two stream 3D-CNNs. The proposed method is able to classify videos based on the learned spatio-temporal networks which does not need optical flow as input and it is benefited from learned optical flow information embedded. We let the network learn the best aspects of optical flow and appearance together in one network by an end-to-end manner using only the RGB frames of the video. We will show that, the mid-level motion features, which are extracted from the trained convolutional neural network can be a good replacement of optical flow with a considerable lower computational cost for video classification. It's shown that the learning motion representation in the way which also is involved in classification problem and inherited action clues performing well in terms of speed and accuracy.

\subsection{Initial Method}
Our first proposed and initial method is to train a 3D Conv-Deconv network to compute optical flow from sequence  of video frames and then combine the mid-level motion features with the RGB 3D-Convnet features to do classification. Figure \ref{fig:1} shows the detail of this initial approach to use a new motion feature with C3D features. Using this method instead of other two stream networks which need optical flow as input has benefits in manner of speed, since it is incomparable faster than those methods. 
However it gains comparable results but the method still needs to be improved. In the next parts we propose our new two stream networks to train both on class label and flow estimation together and address the issues of speed and accuracy.

\subsection{Combined Network} 
The second idea is to train an end-to-end network using 3D-Convnet and 3D-Deconvnet to train on both action class and motion structure together. The 3D-Convnet part is shared between the action classification and flow estimation networks. In an other view the shared 3D-Conv network is the main network with two loss function. The first one is a softmax loss for action classification and the second one is a 3D-Deconv network followed by a voxel-wise loss. So the end-to-end network is providing a more solid solution to perform better than the first method since the new learned representation is optimized to exploit appearance and motion model together obtained from frames and is as fast as C3D method in the test time.

\subsection{3D Conv-Deconv Two-Stream Net}
Our main proposal is to train a 3D two stream network by an end-to-end learning. Figure \ref{fig:3} shows the details of the proposed idea. The appearance stream is an RGB 3d-convnet and the motion estimation stream is a 3d-convnet followed by a 3d-deconvnet which has optical flow as output to learn motion information. As it has been shown in the figure, The softmax loss performs on the concatenation of the last layers of both 3d-convnets. Which it adds the abstract temporal information of the motion stream to the appearance stream to make more rich representation in categorizing of action. 
In the test time, this method performs similar to single frame two-stream nets in accuracy, by 20 times faster frames per second rate.

This single-step training algorithm is beneficial to accuracy of 3D-CNN on videos by exploiting new convolutional features which are shared among the motion learning and action classification tasks. This method also can be considered as a multi-task network for different purposes. Based on some works \cite{multi_task_CNN}, It's proved that network with different sub-tasks can be more efficient and learn stronger feature representation considering all tasks together. 
\\Our method improves the C3D to learn and use better motion representation than knowledge which is just extracted from sequences of RGB frames without using optical flow in training. In this two-stage cascade, optical flow information is stored via training phase and in the test time, there is no need to computer optical flow, so we can classify videos very fast.
 
\subsection{The Networks Architecture}
For our proposed method, we use the networks for 3D-convnet and 3D-deconvnet which are inspired by \cite{c3d,dc3d}. The 3D-convnet has 5 layers of convolution and 2 fully connected layers plus the layer of classes. The filter numbers for each convolution layer are 64, 128, 256, 256 and 256 respectively and 2048 for  fully connected layers. Same as \cite{c3d}, we use filters with size of 3$\times$3$\times$3 for convolutional layers. 3D-deconvnet is the same network as the V2V network in \cite{dc3d} (Refer to the paper for more details).
\section{Experiments}
\label{Experiments}
We have evaluated our proposed networks on the UCF101\cite{UCF101} action video dataset and will try other datasets in future. In this section, first we explain details of the dataset, then results of the experiments and discussion will come.

\subsection{Dataset and Experiments}
The UCF101\cite{UCF101} action dataset have 101 action classes which categories in 5 main types: human-object interaction, body-motion only, human-human interaction, playing musical instruments, and sports. It contains 13320 video clips, split to 3 different sets of train, validation and test data. The reported results will be the average of the accuracy on these 3 sets.   

For training the networks, we use the C3D pre-trained model on the Sports-1M dataset for 3D-Convnet parts and finetune it on the UCF101 dataset. We trained the 3D-Deconv network from scratch by optical flow extracted from UCF101 frames as groundtruth by the Brox's method \cite{brox_opticalFlow}. For evaluation, we train a linear SVM on the extracted features from each network.

\begin{table}
\begin{center}
\begin{tabular}{|l|c|}
\hline
Method & Average Accuracy \\
\hline\hline
C3D (1 net) \cite{c3d} & 82.3 \\
C3D (3 nets) \cite{c3d} & 85.2 \\
Two-stream CNNs\cite{2stream} & 88.0 \\
Very-Deep\cite{verydeep} & 91.4 \\
TDD\cite{TDD} & 90.3 \\
\hline
Ours-Initial & 85.2\\
Ours-Combined net & 87.0\\
Ours-Twostream 3Dnet & 90.2\\
\hline
\end{tabular}
\end{center}
\caption{Comparing the Average Accuracy of our proposed networks with previous works, on the 3 sets of the UCF101 dataset.}
\end{table}

\begin{table}
\begin{center}
\begin{tabular}{|l|c|}
\hline
Method & frames per sec \\
\hline\hline
C3D (1 net) \cite{c3d} & 313 \\
Two-stream CNNs \cite{2stream} & 14.3 \\
iDT+FV \cite{IDT} & 2.1 \\
\hline
Ours-Initial & 210\\
Ours-Combined net & 300\\
Ours-Twostream 3Dnet & 246\\
\hline
\end{tabular}
\end{center}
\caption{Comparing the number of processed frames in one second in our proposed methods with the related works}
\end{table}

\subsection{Results}
In this section we compare the results of the proposed networks with the previous methods in two main factors, accuracy and test time. 
Comparing the mean accuracy of methods has been shown in Table 1. The baseline of the work, which is the appearance stream of our proposed networks, has been reported in the first row of the table (C3D), they also trained 3 different nets and reported the results using these three networks which improved their mean accuracy with about 3 percents. The two stream, very deep two stream and trajectory-pooled deep convolutional descriptors methods are reported in the next rows. The bottom part shows the mean accuracy of our proposed method on the UCF101 dataset. The initial method which has separate training on the appearance stream and action labels and motion stream and optical flows, outperforms the C3D method using a single network, which means that the abstract mid-level feature of the motion stream has the ability of improving the accuracy. Our second proposed network have an end-to-end training phase. With the shared weights which is trained to learn both action and optical flow simultaneously, we expect an improvement on the accuracy, the results support the expectation with 2 percent improvements. Our third proposed network is also an end-to-end network, with more degree of freedom, because of having two separate networks for appearance and motion. The classification of this network performs on the concatenated feature of both layers. The results show that training the networks without weight sharing outperforms the previous proposed networks.

The comparison of the methods in term of speed of the algorithm in frames per second has been made in Table 2. As our proposed methods work without using any optical flow extraction method they have a considerable difference in speed with other methods. Comparing our methods show that the second network with shared weights have the highest speed between our three proposed networks and the other two with two separate networks have a slightly lower speed.

\section{Conclusion}
\label{Conclusion}

We presented novel convolutional neural networks to embed both appearance and motion in human actions video.
Our two-stream 3D network demonstrates efficient scheme to apply 3D-ConvNets and achieve good performance for video classification. We showed the effectiveness of the method in terms of speed and accuracy in run-time. It can obtain accurate results in a speed of very faster than real time.
{\small
\bibliographystyle{plain}
\bibliography{egbib}
}

\end{document}